\pdfoutput=1
\documentclass[11pt]{article}

\usepackage[final]{acl}

\usepackage{times}
\usepackage{latexsym}
\usepackage{amsmath}
\usepackage{graphicx}
\usepackage{amssymb}

\usepackage{multirow}

\usepackage{algorithm}
\usepackage{algpseudocode}
\usepackage{pifont}

\usepackage[most]{tcolorbox}

\usepackage[T1]{fontenc}
\usepackage[utf8]{inputenc}

\usepackage{microtype}

\usepackage{inconsolata}

\usepackage{graphicx}

\title{User Behavior Prediction as a Generic, Robust, Scalable, and Low-Cost Evaluation Strategy for Estimating Generalization in LLMs}

\author{Sougata Saha\textsuperscript{\thanks{Both authors contributed equally to this paper.}} and Monojit Choudhury\textsuperscript{\footnotemark[1]}\\
Mohamed bin Zayed University of Artificial Intelligence,\\
\texttt{\{sougata.saha, monojit.choudhury\}@mbzuai.ac.ae}
}

\begin{document}
\maketitle
\begin{abstract}
Measuring the generalization ability of Large Language Models (LLMs) is challenging due to data contamination. As models grow and computation becomes cheaper, ensuring tasks and test cases are unseen during training phases will become nearly impossible. We argue that knowledge-retrieval and reasoning tasks are not ideal for measuring generalization, as LLMs are not trained for specific tasks. Instead, we propose {\em user behavior prediction}, also a key aspect of {\em personalization}, as a theoretically sound, scalable, and robust alternative. We introduce a novel framework for this approach and test it on movie and music recommendation datasets for GPT-4o, GPT-4o-mini, and Llama-3.1-8B-Instruct. Results align with our framework’s predictions, showing GPT-4o outperforms GPT-4o-mini and Llama, though all models have much room for improvement, especially Llama.
\end{abstract}

\section{Introduction}

\textit{"The central challenge in machine learning is that we must perform well on new, previously unseen inputs—not just those on which our model was trained. The ability to perform well on previously unobserved inputs is called generalization"} - \citet{goodfellow2016deep}.

Utilizing large amounts of data for training, large language models (LLMs) have achieved state-of-the-art performance on existing and new evaluation benchmarks, demonstrating remarkable capabilities over varied use cases. However, increasing training data makes models susceptible to data contamination issues, where the model has been exposed to the test data during training. For example, GPT 3.0 \cite{brown2020language} was exposed to portions of test data, conflating its test scores. Such issues question the efficacy of existing evaluation benchmarks in measuring LLMs' generalizability. Hence, during evaluation, the model can recall from its memory instead of learning the underlying pattern, resulting in conflated performance on existing datasets and a false sense of {\em generalization}. Distinguishing this memorization capacity from learning transferable principles is a key challenge in measuring generalization in foundational models \cite{chu2025sftmemorizesrlgeneralizes}. One might argue that memorization, although a weaker form of generalization, suffices as long as LLMs perform well on tasks of practical importance. However, since the space of all problems is unknown, we do not know in what situations a model might fail. Also, since the world is dynamic, continuous memorization is impractical, which generalization addresses. Thus, not understanding models' generalization capabilities hampers their reliability.%Although there are several popular frameworks and benchmarks for LLM evaluation \cite{chang2024survey} that test for a model's knowledge, reasoning, alignment, and safety properties \cite{guo2023evaluating}, it is unclear how much of these properties are an outcome of generalization and how much can be achieved just through memorization. 

Although several popular frameworks and benchmarks \cite{chang2024survey} for evaluating LLMs' knowledge, reasoning, alignment, and safety properties exist \cite{guo2023evaluating}, it is unclear how much of these properties are due to generalization and how much can be achieved only through memorization. Also, as LLMs' compute capacity increase, it becomes more difficult to create challenging evaluation benchmarks free from novel forms of contamination, such as task contamination \cite{li2024task}, leading to an uptake of complex evaluation benchmarks \cite{ he-etal-2024-olympiadbench} whose practical utility is unknown \cite{Zhou2023DontMY}. Hence, we ask \textit{what should be an ideal strategy for evaluating LLMs' generalization?} The solution must be robust to data contamination; it should be dynamic, and time and cost-efficient, and it must ensure the availability of distinct test sets, even if models utilize \textit{all} available data for training.

%without explicitly capturing the models' generalization capabilities. not understanding their learning dynamics \cite{kang2024learning},

% One might argue that as long as LLM's perform well on tasks of practical importance, generalization does not matter. However, not understanding models' generalization capabilities hampers their reliability as we do not know in what situations a model might fail. 
%We propose measuring LLMs' \textit{personalization} capability as a simple yet robust solution. 
% Hence, we must reflect on the nature of model training. 
% Since most available training data are human-generated, they capture human behavior in a context at a point in time, which LLMs, trained as next-word predictors, intrinsically learn from. Although their evaluation encompasses measures such as perplexity, which locally measures the next word prediction given the current context, inherently learning from human behavior, LLMs should be capable of predicting future behavior in other contexts, which would be a more robust test of its generalization capabilities. 

Since most available and high-quality LLM training data are human-generated, they capture human behavior in a context at a time. Thus, although LLMs are trained as next-word predictors, they should essentially learn the intrinsic task of behavior prediction from context. To this end, we propose {\em user behavior prediction} or \textit{personalization} as a simple yet robust strategy for measuring LLM's generalization over a wide, potentially an infinite, range of capabilities. Although existing studies measure LLMs' personalization capabilities \cite{lin2023can, zhao2024recommender,wu2024survey, dai2023uncovering, liu2023chatgpt}, deviating from the standard definition, we propose a novel framework that uses personalization to measure generalization. By re-purposing existing resources, our method presents a robust and cost-effective measure of generalization in practical settings, which presents a perspective shift in how we utilize them.

In the following section, we argue our position of using personalization to measure generalization and empirically demonstrate an entropy-based framework for measuring generalization. Defining generalization as a model's capacity to follow the actual entropy change with varying context, we first present a statistical framework that evaluates the capacity of existing tasks as good generalization benchmarks and then use recommendation systems as a use case to measure the generalization capabilities of GPT-4o, GPT-4o-mini, and Llama-3.1-8B-Instruct against a baseline. Overall, our contributions are summarized below:
\begin{enumerate}
    \item We propose user behavior prediction, a key aspect of personalization, as a theoretically sound, scalable, and robust alternative to measuring generalization.
    \item We empirically test our hypothesis using an entropy-based framework and present results for movie and music recommendations using GPT-4o, GPT-4o-mini, and Llama-3.1-8B-Instruct.
    \item We discuss the implications of our findings in enabling generalization. %We share our framework for the research community to study.
\end{enumerate}

\section{Generalization in the era of LLMs}
\label{current_issues}
In essence, generalization is a model's ability to perform well on unseen test data, given that the test set measures the same task during model training, indicating the efficacy of the model in learning the underlying patterns in the training data without overfitting or underfitting, attaining a trade-off between bias and variance \cite{bishop2006pattern}. LLMs, conceived as general-purpose models, are expected to follow instructions in natural language and perform well on varied tasks, emulating human-like behavior. However, current task-centric evaluation schemes fail to measure the generalization capacity of models holistically, often leading to conflated results. This deviation is primarily due to the following factors.

\subsection{Issues of Task-Centric Evaluation}

Since training LLMs involves online data, it is crucial to understand the nature of such data. Almost all of the available online data pertain to human behavior. A data point is a user's behavior in a specific context in time, where the context \cite{zimmermann2007operational, bazire2005understanding} characterizes the situation and is the cumulative aggregate of all user behaviors leading to that time. Any training dataset embodies such knowledge and patterns and represents human behavior across time. Hence, although LLM training involves the task of next-word prediction, they are essentially trained on the task of user behavior prediction from the context. Thus, they should be evaluated on similar tasks to meaningfully gauge their generalization capabilities, which task-centric evaluation approaches fail to measure holistically.

Also, unlike other statistical models, LLMs require sizable data for training, which has grown with time. However, since the growth of the total stock of public human text data is asymptotic, LLMs are projected to utilize all available data for training between 2026 and 2032 or even earlier \cite{villalobos2022will, villalobos2024will}, making it difficult to guarantee that the test sets are unseen during training. Although most open-weight LLMs utilize publicly available online text data during pre-training, the exact data splits used during training are unknown. Also, the data used in the fine-tuning and alignment phases is usually private, making it hard to gauge data and task contamination.

Nearly exhausting all available data for training, the probability of data contamination is high in new test sets, which even synthetic approaches to data creation will not mitigate. Hence, we need evaluation frameworks that are novel and free from these issues. Instead of investing in newer datasets, we propose a perspective shift. We propose a framework for measuring generalization by repurposing existing personalization benchmarks, which can be a robust test for generalization, as we shall see. Prior to that lets introduce a formal description of training data.

\subsection{A Formal Description of Training Data}
\label{data_model}

Let $\mathcal{U}$ denote the set of all online users, where $\mathcal{U} = \{u_1...u_n\}$. Let $c_t$ denote the context at any given time $t$, which is an aggregate of all user behavior till time $t-1$. Let $\mathcal{B}^u = \{b^u_1,...,b^u_t\}$ denote all the behavior of user $u$ till time $t$. A data point is a user's behavior $b_t^{u}$ to the context $c_t$ at time $t$, where each user is a function that maps the context to their behavior at a point in time. Psychologists, anthropologists, and linguists often cluster human behavior by variables such as demography. Such variables define the group's behavior and preferences across some dimensions. \citet{adilazuarda2024measuringmodelingculturellms} formally terms such features as demographic \textit{proxies} of culture, which capture the differences of user groups across dimensions termed semantic proxies \cite{thompson2020cultural}. Here, we combinedly refer to these factors as \textit{proxies} and can represent any documented grouping such as geodemography, or undocumented groupings such as {\em dog lovers}. Let $\delta = \{\delta_1...\delta_k\}$ represent the set of all relevant user proxies following a distribution $p_{\delta}$, where $\delta_j \in \delta$ is a grouping of users $\mathcal{U}$. A dataset $\mathcal{D}$ is the set of all triples of contexts, user proxies, and their behaviors, across time. We will refer to this framework in the next sections.

\section{Proposed Method}
\textit{``It is not difficult to devise a paper machine which will play a not very bad game of chess...Are there imaginable digital computers which would do well in the imitation game?"} -\citet{turing2004intelligent}.

Alan Turing observed that a problem is easier when there is an end goal. He argued that machines that can mimic humans in dynamic scenarios are much more intelligent than machines that are good at universal tasks such as playing the game of chess, where the rules of the game are already known \cite{turing1950machinery}. Ludwig Wittgenstein made a much more fundamental observation about language since its rules constantly evolve. Language, as a mode of communication, is meaningful only in the context of the situation, which factors in users and their surroundings. Consider the famous "builder's language" thought experiment \citet{wittgenstein2009philosophical}, where he depicted language as a communication tool in the context of social activity between builder A and assistant B. Builder A is building with blocks, pillars, slabs, and beams. Builder B has to pass the stones in the order specified by A. They use a language consisting of the words "block," "pillar," "slab," and "beam." Builder A calls out an item that Builder B brings. Hence, the builders developed their pragmatic language using only four words, creating what Wittgenstein called a "complete primitive language." The words lose their pragmatic meaning without the context and the builders.

In the current context of LLMs, the ability to perform complex reasoning tasks, such as writing code or solving math problems, although difficult for many humans, are shallower measures of intelligence and generalization since they are universal tasks. However, the ability to solve the same task, mimicking a specific person, is much more challenging than solving the task like any person. Extending Turing's definition, we argue that any machine that can mimic individual users from its training data is the most robust test of intelligence and the strongest measure of generalization. Hence, we propose measuring a model's capacity for personalization as a robust test of its generalization.

\subsection{Personalization as a Test for Generalization}

With the objective of delivering relevant information to an individual or a group of individuals \cite{kim2002personalization}, personalization broadly means tailoring something for an individual without their active participation \cite{fan2006personalization, vesanen2007personalization}. Unlike customization, where individual users are actively involved in tailoring the outcome by specifying their preferences, personalization is without the user's active control and usually involves passively understanding user preferences from their actions \cite{sundar2010personalization}. Considering personalization as a mode of individuation, \citet{lury2019algorithmic} defines it as a recursive approach that "involves forms of de- and re-aggregating, in which a variety of contexts are continually included and excluded" to determine the best possible group affiliation of users. Hence, personalization is a pathway that starts with an initial broad assumption about a user's background, which is constantly refined over time based on their behavior until their optimal preferences are determined \cite{schmitt1999experiential, dhar2000consumer, hanley2006some, droe2006music, wilken2011cultural, rogers2014diffusion, tahmasbi2018modeling}.

The core tenet of digital personalization is that the underlying algorithm should better understand the user's background and preferences with more interactions to facilitate delivering better-personalized content over time. At the risk of anthropomorphizing, LLMs' generalizability is the ability to emulate human-like behavior in real-world applications, which requires understanding human behavior in practical settings. Statistically, it involves generalizing past behavior patterns to predict future behavior, thus embodying Turing's philosophy that intelligent machines should be capable of mimicking human behavior. We argue that any model (algorithm) that can perform well in the personalization task is more generalizable. In the following subsection, we formalize this notion of generalization.

\subsection{A Statistical Framework for Levels of Generalization}
\label{statistical_framework}
% Since data points represent user behavior in a situation (Section \ref{data_model}), any machine learning task ultimately involves behavior prediction given a context, where the number of users in the context might vary from an individual to the entire human population. Borrowing the definition of cultural proxy from \citet{adilazuarda2024measuringmodelingculturellms} to define the user groups, any proxy that can reduce the entropy of a subset more than any random subset of similar size is a meaningful proxy. A model's capability to use such proxies to predict the outcome is a test of its generalizability.

% As discussed in Section \ref{data_model}, any machine learning task ultimately involves group behavior prediction given a context, where the proxy represents the size of the user group, which might vary from an individual to the entire human population. Hence, any proxy that can reduce the entropy of behavior prediction more than any random subset of similar size is a meaningful proxy. A model's capability to use such proxies to predict the outcome is a test of its generalizability.

As discussed in Section \ref{data_model}, since LLMs learn from the entirety of human behavior data, any task ultimately involves group behavior prediction given a context, where the proxy represents the size of the user group, which might vary from an individual to the entire human population. Hence, any proxy that can reduce the entropy of behavior prediction more than any random subset of similar size is a meaningful proxy. A model's capability to use such proxies to predict the outcome is a test of its generalizability.

% LLMs have performed remarkably well on diverse tests, where some models have even attained or surpassed human-level performance on complex knowledge and reasoning-based tasks. Although the difficulty of such tasks varies, the answers are universal and depend primarily on the context, making it easier for models to recall memorized answers from the training set. On the other hand, LLMs perform much worse in cultural tests, where the correct answer varies by culture, making it much more difficult for models to generalize. Generating the correct answer requires generalizing from a group's perspective and factoring in users. Hence, even the simple solution of retrieving answers from their parametric knowledge requires factoring in user information, posing a more difficult task than general knowledge and reasoning-based tasks. Although there is progress in interpreting LLMs' internal memorizing and generalization capacities, it is still difficult to interpret their generation process deterministically. However, benchmarks that require factoring in extraneous information other than just the context are bound to be more difficult. Hence, by treating such models as black boxes and from a behavioral point of view, comparing the model's response between such test sets should indicate its generalizability. To this end, we propose a statistical framework for measuring the levels of generalization.

Given a task $T$, let $\mathcal{B}^T \in \{b_1...b_n\}$ represent the set of all behaviors from all users across time. Let $p(b_i|\delta_j)$ represent the probability of a behavior $b_i \in \mathcal{B}^T$ for proxy $\delta_j \in \delta$. The generalization capacity of a model $\theta$ for the task is inversely proportional to the expected difference between the cross-entropy $\hat{H(\mathcal{B}^T_{\delta_j})}$ and the true entropy $H(\mathcal{B}^T_{\delta_j})$ for each proxy $\delta_j$, as defined below:

\begin{align}
   & H(\mathcal{B}^T_{\delta_j}) = -\sum_{b_i \in \mathcal{B}^T} p(b_i|\delta_j)\log  p(b_i|\delta_j) \\
   & \hat{H(\mathcal{B}^T_{\delta_j})} = -\sum_{b_i \in \mathcal{B}^T} p(b_i|\delta_j)\log  p(b_i|\delta_j, \theta)
   % & G_{\theta}^T \propto \frac{1}{\mathbb{E}_{\delta_j \sim p_{\delta}} [\hat{H(\mathcal{B}^T_{\delta_j})} - H(\mathcal{B}^T_{\delta_j})] + \epsilon}
\end{align}
% Ideally, $H(\mathcal{B}^T_{\delta_i})$ and the $\hat{H(\mathcal{B}^T_{\delta_i})}$ should be equal, indicating best generalization for task $T$.

Depending on the dependence of the behavior and the proxy, the notional complexity of a task is a continuum from weak to strong as below:

% As depicted in Equation \ref{ml_dataset}, since tasks embody user behavior, their complexity depends on the strength of the relationship between the behavior and the user proxy. Hence, the notional complexity of a task is a continuum from weak to strong, depending on the reliance on proxies to solve the task as depicted below:

\noindent
\textbf{Weakest Case} ($\mathcal{B}^T \!\perp\!\!\!\perp \mathcal{U} | c_t)$: Tasks where the behavior depends on the context $c_t$ and is independent of individual users $\mathcal{U}$ or their proxies $\delta$, making them notionally less complex. Most knowledge and reasoning-based tasks such as MMLU \cite{hendrycks2020measuring}, GSM8K \cite{cobbe2021training}, GLUE \cite{wang2018glue}, etc, are examples of such kinds, which measure universal patterns independent of proxies, hence notionally weakest test sets for generalizability.

\noindent
\textbf{Average Case} ($\mathcal{B}^T \!\perp\!\!\!\perp \mathcal{U} | c_t, \delta_j$, where $ \delta_j \sim p_{\delta}$): Tasks where the outcome is independent of individual users but dependent on their proxies and the context are notionally more complex. For example, cultural evaluation benchmarks that require group-specific reasoning are stronger test sets for generalizability \cite{li2024culture, alkhamissi-etal-2024-investigating, nadeem-etal-2021-stereoset,nangia-etal-2020-crows, wan-etal-2023-personalized, jha-etal-2023-seegull, li2024culturegenrevealingglobalcultural, cao-etal-2023-assessing, tanmay2023probingmoraldevelopmentlarge, rao-etal-2023-ethical, kovač2023largelanguagemodelssuperpositions}.

\noindent
\textbf{Strongest Case} ($\mathcal{B}^T \not\!\perp\!\!\!\perp \mathcal{U} | c_t, \delta_j$, where $ \delta_j \sim p_{\delta}$): Tasks where the outcome depends on individual users are notionally most complex. For example, user-specific tasks, such as item recommendation, necessitate reasoning from a user's perspective and are notionally more complex and best evaluation benchmarks for generalizability \cite{nagarnaik2015survey, ko2022survey}.

% On one end are most knowledge and reasoning-based tasks, where the expected outcome is universal and independent of user proxies, making them notionally less complex. Evaluation sets such as MMLU, GSM8K, GLUE, etc, are examples of such kinds, which measure universal understanding independent of proxies, hence notionally less complex. On the contrary, user-specific tasks, such as item recommendation, necessitate reasoning from a user's perspective and are notionally more complex. Any other tasks that rely on varying levels of user proxies, such as cultural reasoning benchmarks, lie in between in terms of notional complexity. 

% $\mathcal{B}^T \!\perp\!\!\!\perp \mathcal{U} | c_t, \delta$
% $\mathcal{B}^T \!\perp\!\!\!\perp \mathcal{U} | c_t, \delta_j$, where $ \delta_j \sim p_{\delta}$
% $\mathcal{B}^T \not\!\perp\!\!\!\perp \mathcal{U} | c_t, \delta_j$, where $ \delta_j \sim p_{\delta}$
\subsection{Hypothesis}
\label{hypothesis}

\begin{figure}[h]
    \centering 
    \includegraphics[width=\columnwidth]{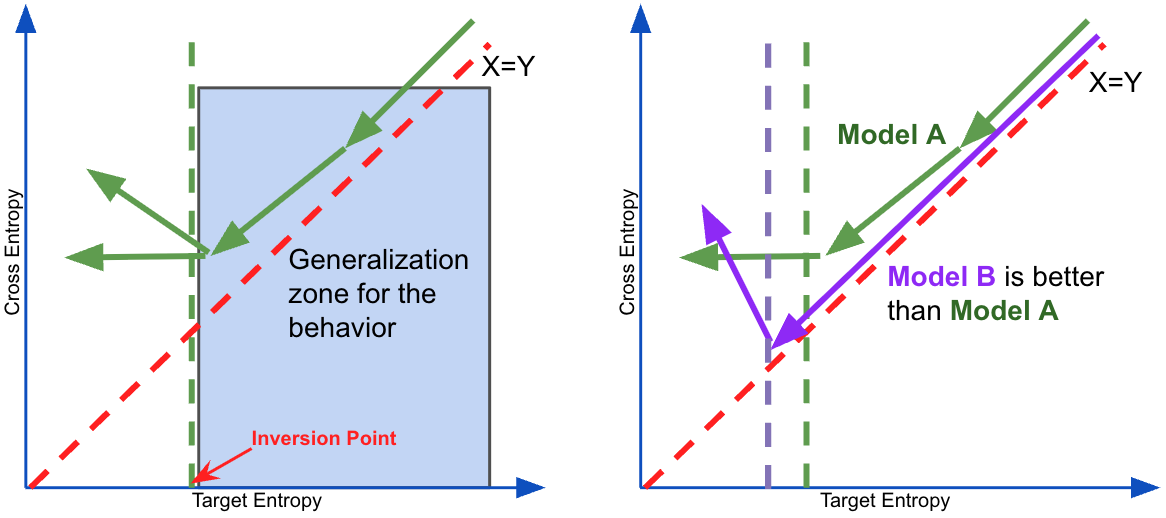} 
    \caption{(Left) Hypothesized Behavior. (Right) Hypothesized Model Comparisons.}
    \label{fig:ideal_graphs_hypothesis}
\end{figure}

\noindent
As depicted in Figure \ref{fig:ideal_graphs_hypothesis} (left), ideally, the distribution of the target $H(\mathcal{B}^T_{\delta_i})$ and cross-entropy $\hat{H(\mathcal{B}^T_{\delta_i})}$ should be equal. When for a model tested under different cases the points ($H(\mathcal{B}^T_{\delta_i})$, $\hat{H(\mathcal{B}^T_{\delta_i})}$) are plotted on a graph, in the ideal case of generalization, $H(\mathcal{B}^T_{\delta_i}) = \hat{H(\mathcal{B}^T_{\delta_i})}$. Or in other words, the points should lie on the X=Y line. However, in reality, we expect $H(\mathcal{B}^T_{\delta_i}) <  \hat{H(\mathcal{B}^T_{\delta_i})}$. This gap is expected to be small when $H(\mathcal{B}^T_{\delta_i})$ is high (i.e., the average case of generalization when only proxies are used to predict behavior). We expect, therefore, the plot to follow the X=Y line for large X, but then flatten out or even rise for lower X. The point at which this inversion of behavior happens is the point when the model can no longer generalize to specific users' or groups' behavior. As depicted in Figure \ref{fig:ideal_graphs_hypothesis} (right), we hypothesize the inversion point to change across models, where a lower inversion point indicates a better generalizable model. Where do current LLMs lie in this framework of generalization? We put our hypothesized statistical framework to test in the next sections.

\section{Measuring Generalization via Personalization}
\label{measuring_generalization}

Recommendation systems, at their core, are personalization engines. They are predictors of a user's future behavior based on some observed past behavior. We test our proposed statistical framework in Section \ref{statistical_framework} and experiment with movie and music recommendations, where the task is to recommend a list of N items based on the user's history.

\subsection{Dataset and Preprocessing}
We experiment with the MovieLens\footnote{\url{https://grouplens.org/datasets/movielens/1m/}} \cite{harper2015movielens} and last.fm\footnote{\url{http://last.fm/}} \cite{Celma:Springer2010} datasets since both these datasets contain demographic information and are widely used in recommendation systems literature. Also, since recommendation datasets are known to be very sparse, we preprocess both datasets to reduce sparsity. 
Collected by GroupLens Research, the MovieLens dataset (movies dataset) contains 1 million ratings of approximately 3,900 movies made by 6,040 MovieLens users, along with their demographic information such as gender, age, and occupation. This is a well maintained dataset and extensively used in recommendation systems \cite{goyani2020review} literature. As a post processing step, we remove users with occupation listed as `others' and restrict to demographic groups (combination of age, gender and occupation) containing at least 30 users.

Containing the music listening habits of nearly 1,000 users, along with their demographic information such as gender, age, and country, the last.fm dataset (music dataset) is widely used in music recommendation literature \cite{schedl2016lfm}. It has also impacted research pertaining to music and mood \cite{ccano2017music}, and other cultural and behavioral studies \cite{chen2010last, putzke2014cross}. However, the number of users from each country follows a long-tailed distribution. Hence, as a post processing step, we derive the continent proxy based on country and restrict to users from Europe, North America, South America, and United Kingdom who have at least 5,000 interactions.

% \noindent
% \textbf{Preprocessing:} Since recommendation datasets are known to be very sparse, we preprocess both datasets to reduce sparsity. Although the music dataset has country-level information, the number of users from each country follows a long-tailed distribution. Hence, we derive the continent proxy based on country and restrict to users from Europe, North America, South America, and United Kingdom who have at least 5,000 interactions. For movies we remove users with occupation listed as `others' and restrict to demographic groups (combination of age, gender and occupation) containing at least 30 users.

\begin{figure*}[!t]
    \centering 
    \includegraphics[width=\linewidth]{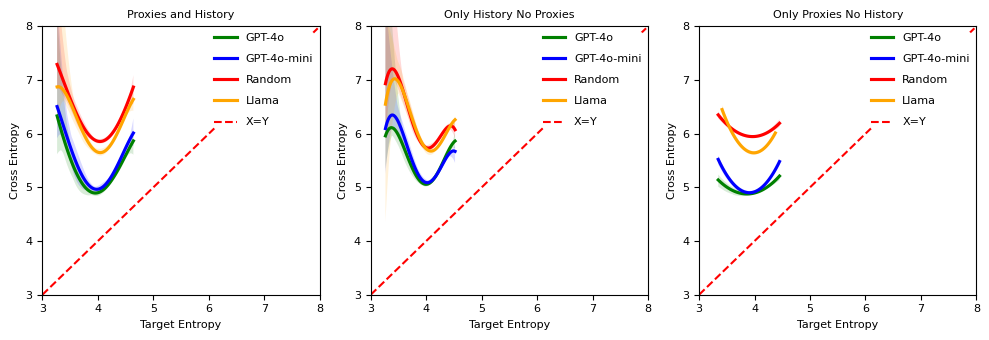} 
    \caption{Movie Entropy Trend. Setup A: Left, B: Middle, C: Right.}
    \label{fig:movie_entropy_trend}
\end{figure*}

\begin{figure*}[!t]
    \centering 
    \includegraphics[width=\linewidth]{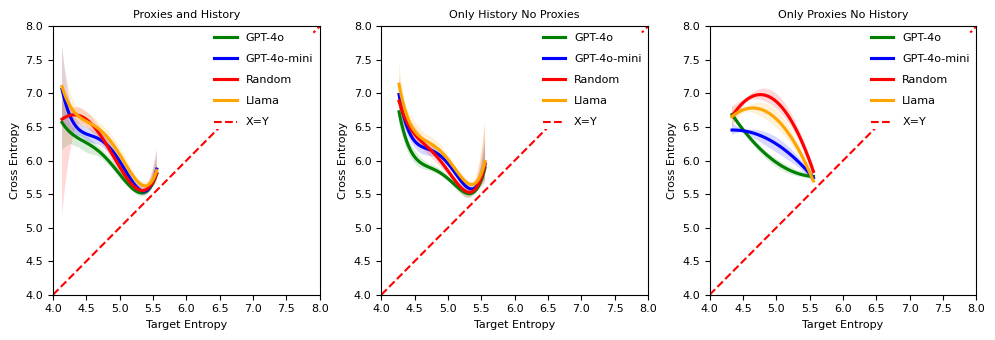} 
    \caption{Music Entropy Trend. Setup A: Left, B: Middle, C: Right.}
    \label{fig:music_entropy_trend}
\end{figure*}

\subsection{Setup}

\noindent
% \textbf{Objective}: A proxy is any feature that enforces a grouping of users, as defined in Section 2.2. Hence, demography and user history are proxies since they can group users based on their background, viewing history, or combinations of both. Also, not all proxies are equal. Any proxy that can reduce the entropy of the outcome more than any random subset of similar size is a much more meaningful proxy for behavior prediction. Our objective is to \textit{measure a model's generalization capacity at different levels of proxies.}

\noindent
\textbf{Experiments:} We set up the experiments as a behavior prediction task and experiment with the following three levels of proxies: 

\noindent
\textbf{(A) Demography and history:} We provide demography $D$ and prior interactions $h$ as the proxy in the context. For example, "recommend 10 movies from the candidates $C$ for 25-30-year-old self-employed females who have watched the Titanic and Pather Panchali." This is equivalent to the \textbf{Average Case} of measuring generalization, as discussed in Section \ref{statistical_framework}.

\noindent
\textbf{(B) Only history:} We only provide prior interactions as the proxy in the context. The intended subgroup is users who have interacted with at least 60\% of the items in the history. For example, "recommend 10 music from the candidates for users who have listened to Bohemian Rhapsody and Comfortably Numb." This is equivalent to the \textbf{Strongest Case} of measuring generalization.

\noindent
\textbf{(C) Only demography:} This is the default and the \textbf{Weakest Case} of measuring generalization case where we solely provide the demographic proxies as context. The intended subgroup is all users from the demography. For example, "recommend 10 movies from the candidates for 25-30-year-old self-employed females."

We experiment with varying lengths of history, such as 0, 1, 3, 5, 10, and 20, and also intersections of demographic proxies, such as combining age, gender, and occupation for movies and gender and continent for music. For each setup, we follow Algorithm \ref{candidate_selection_history} to sample the candidate items. The target distribution is the aggregated probability of a subset of the un-interacted items for all users defined by the proxy ($C_2$), along with items the group will never interact with ($C_1$). For Setup B we input $D = \{\emptyset\}$ as the demographic proxy.

% We set up the experiments as a behavior prediction task and experiment with the following three levels of proxies: \textit{(i) Only demography:} This is the default case where we solely provide the demographic proxies as context. \textit{(ii) Only history:} We only provide prior interactions as the proxy in the context. \textit{(iii) Demography and history:} We provide demography and prior interactions as the proxy in the context. We experiment with varying lengths of history such as 0, 1, 3, 5, 10, and 20, and also intersections of demographic proxies such as combining age, gender, and occupation for movies, and gender and continent for music.

\noindent
\textbf{Prompts:} 
In each setup, we prompt models to recommend a ranked list of 10 items from a candidate list of 50 items for the users defined by the specified proxies in the context. Appendix \ref{fig:sample_prompt} depicts a sample prompt from the movie domain. %We follow Algorithm \ref{candidate_selection_history} to sample the candidates and determine the target distributions for each setup. For case (ii) we input $D = \{\emptyset\}$ as the demographic proxy. 
Our test set comprises approximately 5,000 prompts for each domain. Table \ref{tab:freq_dist_examples} shows the distribution of prompts for each setting in both domains.

\noindent
\textbf{Models:} 
We conducted experiments using GPT-4o \cite{achiam2023gpt}, GPT-4o-mini, Llama-3.1-8B-Instruct \cite{dubey2024llama}, and a random baseline model where 10 recommended items were selected randomly from a pool of 50 candidates. For all experiments, the temperature parameter was set to 0. Running each combination (levels of proxy) incurred a cost of approximately USD 30 for GPT-4o and USD 2 for GPT-4o-mini. Thus, the total expenditure for GPT-based experiments was approximately 6 x USD 32 = USD 192. Llama experiments, on the other hand, were executed on two 48 GB NVIDIA RTX 6000 Ada GPUs, requiring around 18 hours to complete all settings across both domains.

\begin{algorithm}
\caption{Candidate selection algorithm}
\label{candidate_selection_history}
\begin{algorithmic}[1]
\Procedure{Candidate Selection}{}
\State Input: Inventory $I$, Demography $D$, $K$=50
\For{history $h \in \{0, 1, 3, 5, 10, 20\}$ }

\If{$h > 0$}
    \State $I_h$: Sample $h$ random items from $I$
    \State $u_h$: Users with >= 60\% of $I_h$ + $D$.
    \If{|$u_h$| < 3}
        \State Break
    \Else
        \State Continue
    \EndIf
\Else
    \State $u_h$:  Users with $D$.
\EndIf
\State $I_h^u$ = Set of all items interacted by $u_h$.

\State $C_1$ = $K/2$ random items from  $I_h^{uC}$ 
\State $C_2$ = $K$-|$C_1$| random items from $I_h^u$-$I_h$
\State Candidates = Random shuffle $C_1 + C_2$
\State Target distribution: $Freq(\text{Candidates})$
\EndFor
\EndProcedure
\end{algorithmic}
\end{algorithm}

\noindent
\textbf{Evaluation: } Since we only prompt the model to generate a ranked list of 10 items, we approximate the prediction distribution over 50 items by imposing the ground distribution. We sort the target probabilities in descending order and assign the top 10 probability scores to the model's prediction. The remaining 40 items, which are not in the model's prediction, are sorted in descending order of their target probabilities and assigned the remainder of the target probabilities. Thus providing an optimistic estimate of the model predictions.

To test our hypothesis from Section \ref{hypothesis}, we calculate the cross-entropy between the target probability distribution and the model's estimated distribution and plot the results. 
%Ideally, the distribution of the target and cross-entropy should lie on the X=Y line for an ideal generalizable model. The further the distribution, the less generalizable the model. 
To smooth the graphs, we bucket the target entropy in 200 bins and average the cross-entropy score at each interval. We calculate a rolling average of the cross-entropy scores with a window length of 30 and fit an order four polynomial regression curve to visualize the \textit{inflection point}.

\begin{figure*}[!t]
    \centering 
    \includegraphics[width=\linewidth]{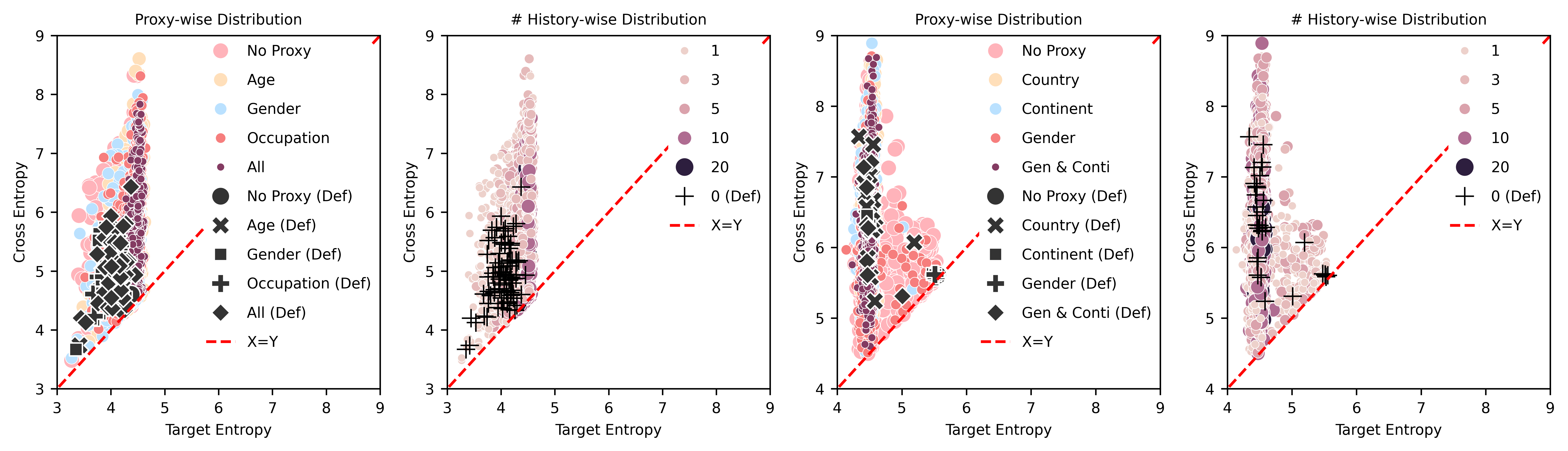} 
    \caption{GPT-4o Demographic Proxy and History-wise distributions for Movie (Left 2) and Music (Right 2)}
    \label{fig:gpt-4o-overall-both}
\end{figure*}

\begin{figure*}[!t]
    \centering 
    \includegraphics[width=\linewidth]{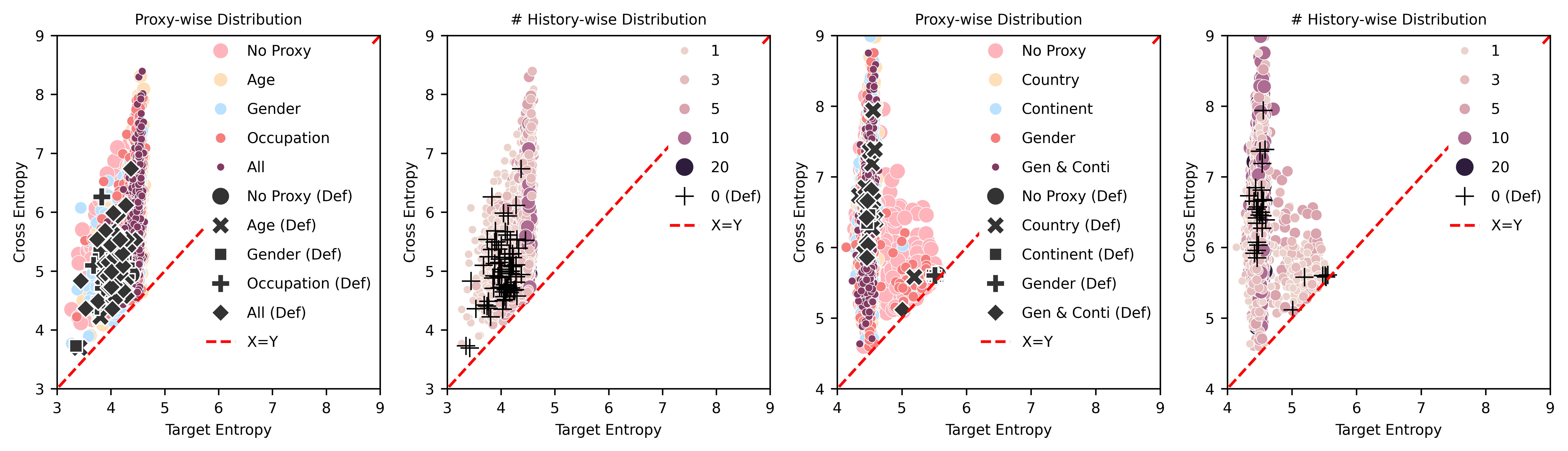} 
    \caption{GPT-4o-mini Demographic Proxy and History-wise distributions for Movie (Left 2) and Music (Right 2)}
    \label{fig:gpt-4o-mini-overall-both}
\end{figure*}

\begin{figure*}[!t]
    \centering 
    \includegraphics[width=\linewidth]{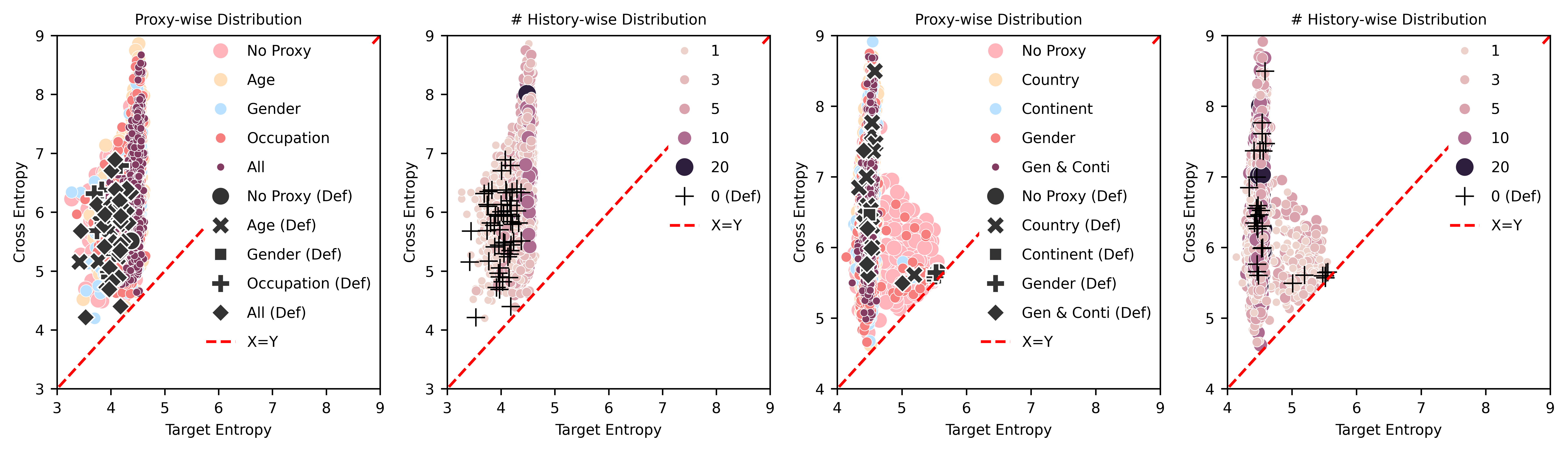} 
    \caption{Llama Demographic Proxy and History-wise distributions for Movie (Left 2) and Music (Right 2)}
    \label{fig:llama-overall-both}
\end{figure*}

\begin{figure*}[!t]
    \centering 
    \includegraphics[width=\linewidth]{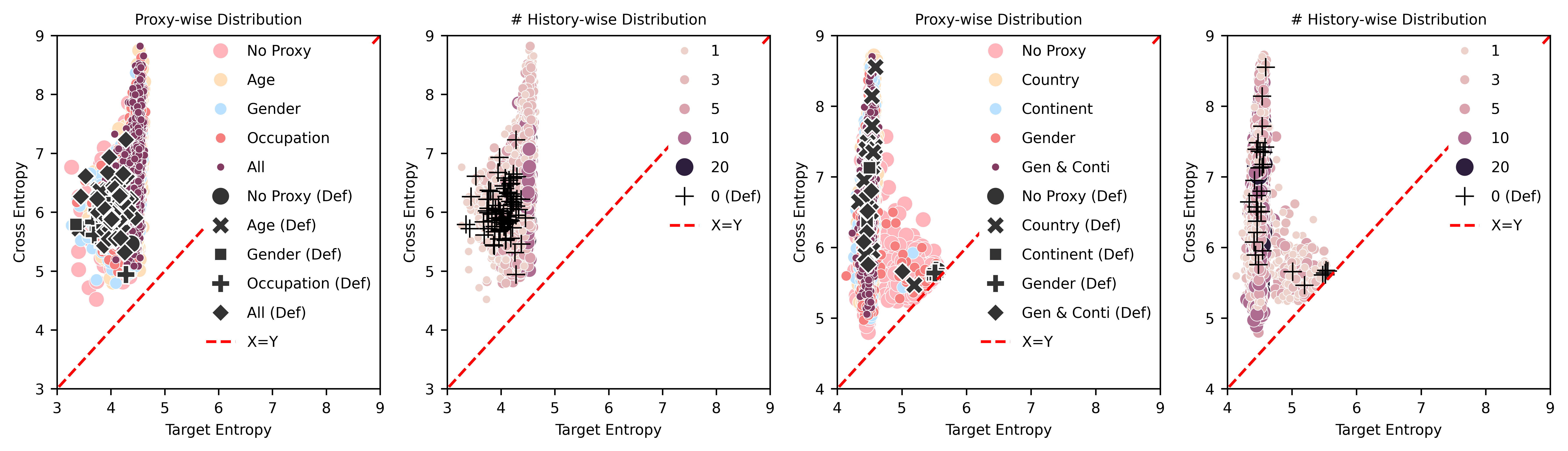} 
    \caption{Random Demographic Proxy and History-wise distributions for Movie (Left 2) and Music (Right 2)}
    \label{fig:random-overall-both}
\end{figure*}

\subsection{Results and Observations}

\textbf{Trends of Generalization}
Figure \ref{fig:movie_entropy_trend} plots the regression curve for the \textbf{movies} domain. We observe the following: (i) The inflection point in all three setups is much lower for LLMs than the random baseline, indicating that \textit{models generalize to a certain degree}. However, they widely differ in absolute numbers, where Llama performs much worse than GPT-4o and GPT-4o-mini, and slightly better than the random baseline. Both the curves for the GPT-based models are closer to the ideal X=Y line, and \textit{GPT-4o slightly outperforms GPT-4o-mini} in all setups, with a lower inflection point. (ii) Since a lower target entropy signifies fewer users in a proxy, which necessitates more specific predictions, an increase in cross-entropy with decreasing target entropy indicates that the \textit{models are not capable of personalizing predictions to smaller subsets of users, and hence will not be good at modeling each individual.} This is also evident from Setup C, which only prompts using demography and no history, and hence the weakest test of generalization. Almost following the X=Y line, the inflection points in all models are much lower than in other setups, signifying that \textit{models can generalize using broader proxies and fail when the number of representative users in a proxy decreases}. (iii) The inflection point is lower in setup A than B, signifying that \textit{combining demography and history enables model prediction}. This indicates that providing cultural information can enable personalization-based tasks.

% Figure \ref{fig:movie_entropy_trend} plots the regression curve for the \textbf{movies} domain. We observe the following: (i) The inflection point in all three setups is much lower for GPT-4o and GPT-4o-mini compared to the random baseline, indicating that \textit{models generalize to a certain degree}. Both the curves for the GPT-based models are closer to the ideal X=Y line. However, \textit{GPT-4o slightly outperforms GPT-4o-mini} in all setups, with a lower inflection point. (ii) Since a lower target entropy signifies fewer users in a proxy, which necessitates more specific predictions, an increase in cross-entropy with decreasing target entropy indicates that the \textit{models are not capable of personalizing predictions to smaller subsets of users, and hence will not be good at modeling each individual.} This is also evident from Setup C, which only prompts using demography and no history, and hence the weakest test of generalization. Almost following the X=Y line, the inflection points in both models are much lower than in other setups, signifying that \textit{models can generalize using broader proxies and fail when the number of representative users in a proxy decreases}. (iii) The inflection point is lower in setup A than B, signifying that \textit{combining demography and history enables model prediction}. This indicates that providing cultural information can enable personalization-based tasks.

Figure \ref{fig:music_entropy_trend} plots the regression curve for the \textbf{music} domain. We observe the following: (i) All \textit{models perform much worse than in movies}. Their inflection points are much higher. The predictions deviate from the X=Y line in all three setups, signifying the model's incapability of generalization using the proxies. (ii) \textit{GPT-4o performs best} in all setups, where Llama is closer to the random baseline. This signifies that \textit{music prediction is inherently more difficult than movie prediction} since the music dataset contains more interactions than movies, indicating that people listen to more music than watch movies. This can also signify that \textit{the set of proxies used in the experiment is not optimal for personalizing music}. (iii) The inflection point of GPT-4o in Setup C is lower, indicating its generalization capabilities using broader proxies. Also, similar to movies, the inflection point of all models is lower in setup A than B, signifying that \textit{combining demography and history enables model prediction.} 

\noindent
\textbf{Overall and Proxy-wise distributions}

\noindent
We also plot the demographic and history-wise distributions of the cross-entropy in both domains in Figures \ref{fig:gpt-4o-overall-both} to \ref{fig:random-overall-both}. The term "(Def)" represents the default setting where no prior history is provided. In the left-most and second-from-right plots of the figures, this default setting is further broken down by demographic proxies. It reflects the model's responses when tasked with recommending the next 10 music or movie items from a given list, without being influenced by any prior context. This setup aims to capture the model's intrinsic tendency to associate items with specific cultural proxies. For the right-most and second-from-left plots, "(Def)" indicates a scenario where the model is probed using all proxies combined but still without any history. In this case, the model is presented with a list of music or movie items and asked to predict the next 10 items solely based on the given list, without any prior history or proxy influence. This configuration is designed to assess the model's inherent affinity for items within the provided list.

Plotting the demographic proxy-wise distribution, in Figures \ref{fig:gpt-4o-overall-both}, \ref{fig:gpt-4o-mini-overall-both}, and \ref{fig:llama-overall-both} for GPT-4o, GPT-4o-mini, and Llama, we observe that the \textit{models are incapable of generalizing when the combination of demographic proxies increase, irrespective of the size of history}, which is the \textbf{Strongest Case} for measuring generalization, according to our proposed framework in Section \ref{statistical_framework}. For example, in movies (leftmost chart), the cross-entropy increases when a combination of all proxies is used along with history (Proxy = All). We see a similar trend in music (second from right), where although the target entropy decreases with more proxies (Proxy = Gen \& Conti), the cross-entropy increases.

In both domains (second from left and rightmost), we see models exhibiting a similar behavior with different lengths of history, where the cross-entropy increases with more history, which is the \textbf{Strongest Case} for measuring generalization. This indicates the model's incapability to adeptly utilize the context, which is a known phenomenon \cite{mukherjee-etal-2024-cultural}.

We further plot the results of each demographic proxy and history independent in Section \ref{additional_plots} (Appendix \ref{sec:appendix}). Overall, our experiments indicate the viability of our proposed framework in Section \ref{statistical_framework} for measuring generalization using existing personalization-based tasks. We clearly see that models are capable of generalization to a certain degree till they reach an inflection point. Although GPT-4o and GPT-4o-mini models perform better than Llama and a random baseline, the results indicate a strong generalization gap and much room for improvement.

\section{Discussion}
\noindent
\textbf{1. What is the difference between personalization and personalization as a measure of generalization?} Personalization as a measure of generalization is a theoretical framework for measuring generalization through the lens of personalization, not on improving personalization itself. Unlike existing personalization techniques \cite{sun2023measuring, hwang2023aligning, zhang2024personalization}, which are evaluated using accuracy-based metrics (e.g., F1, NDCG, MAP), our method analyzes generalization by examining cross-entropy over the model's response distribution. This fundamental difference in objectives and metrics makes direct comparisons with prior works inherently challenging.

\noindent
\textbf{2. Why is personalization interesting and useful for studying generalization in LLMs?}
Besides the theoretical arguments provided earlier, here we list some compelling practical reasons for favoring personalization as an evaluation strategy for generalization.
% Our arguments for personalization are as follows:

\noindent
\textbf{Generalizing from learned knowledge:} Since the context length of current models is limited, personalization requires reasoning using more context, which might be outside the model's context length. Thus, it is a robust measure of a model's generalizability as its capacity to leverage the learned world knowledge during training.
%For example, the collaborative way of recommendation is a 2-step process. It first requires establishing user groups by grouping similar users and aggregating their behavior as the estimated group behavior for items. The second step is assigning the current user to a group based on their behavior and recommending the most prevalent other behaviors of the group members, which are novel for the user, as the recommended item or behavior. The first step necessitates learning an appropriate worldview during training. The second step requires generalizing that knowledge during inference in real-world settings.

\noindent
\textbf{Balancing worldviews:} Since personalization requires tailoring things for an individual, a generalizable model should be capable of balancing between universal and individual-specific knowledge for performing tasks. 
% Although collaborative methods are plausible for new users when their individual preferences are unknown, personalization necessitates improving the estimate of the user's preferences based on their subsequent interactions to recommend relevant items. Hence, with more interactions, personalization requires a shift in perspective from a general worldview to a user-centric worldview. As with the elephant's example, a model should be capable of using each individual's interactions as the frame of reference to establish their worldview and act accordingly. Measuring the capacity of this shift from a general to an individual frame of reference is a more robust measure of generalization.

\noindent
\textbf{Dynamicity:} Personalization evades the issue of models memorizing training examples and recalling during inference, since individual preferences change over time. Hence, a model can't blatantly memorize each user's behavior to perform well in personalization tasks.% It requires learning the core tenet of personalization and generalizing. Although memorization is a form of generalization, it is a weaker measure than personalization or any other reasoning task.

\noindent
\textbf{Cost efficiency and ROI:} Our evaluation framework is highly cost-effective compared to creating entirely new benchmarks from scratch. Developing "challenging" test beds for model generalization requires substantial human and computational resources to resolve data scarcity and contamination issues (see Section \ref{current_issues}). In contrast, our approach repurposes existing personalization tasks to assess generalization, eliminating the need for costly new dataset creation. %Since LLMs nearly exhaust all data during training, it becomes costly to create newer datasets that robustly measure generalization. Using our framework of personalization for measuring generalization provides a new perspective that does not require additional cost.
% Although LLMs as recommendation systems are already tested, we propose a framework to measure generalization using the existing recommendation tasks, making it a cheaper and more robust alternative. Due to the increasing training data consumption in LLMs, creating novel evaluation benchmarks free from data contamination is difficult and costly. Although novel test sets such as question answering on unseen examples require learning reasoning patterns from a larger context and applying them to the current example, it is costly to meticulously create test sets that ensure the new data is indeed unseen and free of data contamination.

\noindent
\textbf{LLMs as world models:} Agentic models \cite{shavit2023practices, acharya2025agentic} require universal knowledge, which is less dependent on individual users. Although LLMs have exhibited tremendous capacity as agents, they must model each individual to be world models.

\section{Conclusion}
We propose a statistically motivated framework using personalization to assess generalization in LLMs. Since LLMs are trained on vast human-generated data, we argue that true generalization lies in predicting human behavior rather than specific tasks. This philosophically aligns with Wittgenstein's language games and Turing's imitation game, though with a key distinction with the latter: while Turing’s test requires mimicking {\em any human}, our framework challenges models to replicate {\em a specific user} at varying complexity levels. This shift has profound philosophical and mathematical implications, only some of which we could explore in this paper.
%In this work, we presented a statistically motivated framework for using personalization as a strategy for estimating generalization of LLMs. The fundamental assumption of our framework is that since LLMs are trained on almost all data generated by human over a long period of time, ideally, the test of generalization should be to predict behavior of humans rather than specific tasks that might already be present in the dataset. This argument aligns well with Wittgenstein's notion of language games and Alan Turing's imitation game as a test of intelligence. Though we would like to emphasize that in the original version of the imitation game, the system has to make the human judge wrongly believe that it were {\em a human}, that is to say, {\em any user}. However, our framework demands the model to mimic a {\em particular user}, with varying degrees of complexities. This shift of perspective has deeper philosophical as well as mathematical implications, only some of which we were able to tease apart in this paper.

We would like to highlight the lack of datasets where complex user behaviors or preferences are available alongside their demographic proxies, which will enable us to conduct large-scale and more extensive studies of generalization.

\section*{Limitations}
The empirical study presented here is limited in several ways: First, we explore only two kinds of behavior preferences - movies and music; the choices were motivated by the availability of large public datasets of user preferences or behavior, where we have some demographic proxies for the users. These experiments do not tell us how models generalize for more complex user behaviors or other kinds of demographic proxies (including psychological features such as personality traits). Second, we experiment only with English prompts and Latin script. It will be interesting to compare a model's generalization in the English language and Latin script to that of other languages and scripts, especially when the prompt is expressed in those languages/scripts. Third, our experiments only consider three models - GPT4-o, GPT4-o-mini, and Llama-3.1-8B-Instruct. Since both of the GPT models are closed-source and behind a pay wall, we are aware that reproducing our experiments independently would incur additional cost. Extending the study to more open-weight models, apart from Llama, would be important to understand the robustness of the proposed framework.

Our theoretical framework assumes that it is possible to estimate the true distributions of user behavior from large samples, which might not be the case if user behavior is non-stochastic or chaotic. Furthermore, the datasets used in this study may not be large enough for estimating user behavior at a global or national scale, which implies that our estimates might have large noise terms, leading to significant over- or under-estimations of models' generalizability.

\section*{Ethical Implications}
Being a theoretical and exploratory study, our work has no direct risks or harms. Nevertheless, we assume in our work that the behavior of users can be estimated in a statistical sense for groups of different sizes, and for a certain definition of groups, the entropy of these distributions is small. It is possible to misinterpret this assumption as a promotion of the idea of stereotypical behaviors of certain groups. We warn against such interpretations. The only two assumptions made here are (a) user behaviors can be stochastically modeled (a common assumption made across many branches of social sciences, such as Economics and Psychology), and therefore, (b) there are latent variables that determine such behaviors. Although we have used ``demographic proxies" as a term for these latent variables and used certain proxies (country, age, gender, etc.) in our experiments, we do not promote the idea that users from the same demographic group display similar behavior. The terminology is borrowed from previous work~\cite{adilazuarda2024measuringmodelingculturellms}; however, the proposed framework is agnostic to any anthropological or psychological theory of human behavior.

\section*{Acknowledgements}
This research was supported by Microsoft
Accelerate Foundation Models Research (AFMR) Grant.

% Bibliography entries for the entire Anthology, followed by custom entries
%\bibliography{anthology,custom}
% Custom bibliography entries only
\bibliography{custom, acl_latex}

\clearpage
\appendix

\section{Appendix}
\label{sec:appendix}

\begin{table*}[]
\centering
\resizebox{.85\columnwidth}{!}{%
\begin{tabular}{|c|c|c|c|}
\hline
\textbf{Domain}         & \textbf{Setting}              & \textbf{\# History} & \textbf{Freq} \\ \hline
\multirow{29}{*}{Movie} & No Proxy (Def)                & 0                   & 1             \\ \cline{2-4} 
                        & \multirow{5}{*}{No Proxy}     & 1                   & 300           \\
                        &                               & 3                   & 300           \\
                        &                               & 5                   & 300           \\
                        &                               & 10                  & 89            \\
                        &                               & 20                  & 5             \\ \cline{2-4} 
                        & Age (Def)                     & 0                   & 7             \\ \cline{2-4} 
                        & \multirow{5}{*}{Age}          & 1                   & 300           \\
                        &                               & 3                   & 300           \\
                        &                               & 5                   & 300           \\
                        &                               & 10                  & 147           \\
                        &                               & 20                  & 5             \\ \cline{2-4} 
                        & Gender (Def)                  & 0                   & 2             \\ \cline{2-4} 
                        & \multirow{5}{*}{Gender}       & 1                   & 300           \\
                        &                               & 3                   & 300           \\
                        &                               & 5                   & 300           \\
                        &                               & 10                  & 110           \\
                        &                               & 20                  & 5             \\ \cline{2-4} 
                        & Occupation (Def)              & 0                   & 17            \\ \cline{2-4} 
                        & \multirow{5}{*}{Occupation}   & 1                   & 300           \\
                        &                               & 3                   & 300           \\
                        &                               & 5                   & 300           \\
                        &                               & 10                  & 112           \\
                        &                               & 20                  & 2             \\ \cline{2-4} 
                        & All (Def)                     & 0                   & 51            \\ \cline{2-4} 
                        & \multirow{4}{*}{All}          & 1                   & 300           \\
                        &                               & 3                   & 300           \\
                        &                               & 5                   & 300           \\
                        &                               & 10                  & 43            \\ \hline
\multirow{30}{*}{Music} & No Proxy (Def)                & 0                   & 1             \\ \cline{2-4} 
                        & \multirow{5}{*}{No Proxy}     & 1                   & 246           \\
                        &                               & 3                   & 249           \\
                        &                               & 5                   & 246           \\
                        &                               & 10                  & 240           \\
                        &                               & 20                  & 11            \\ \cline{2-4} 
                        & Country (Def)                 & 0                   & 17            \\ \cline{2-4} 
                        & \multirow{5}{*}{Country}      & 1                   & 250           \\
                        &                               & 3                   & 249           \\
                        &                               & 5                   & 248           \\
                        &                               & 10                  & 208           \\
                        &                               & 20                  & 9             \\ \cline{2-4} 
                        & Continent (Def)               & 0                   & 4             \\ \cline{2-4} 
                        & \multirow{5}{*}{Continent}    & 1                   & 250           \\
                        &                               & 3                   & 249           \\
                        &                               & 5                   & 249           \\
                        &                               & 10                  & 249           \\
                        &                               & 20                  & 11            \\ \cline{2-4} 
                        & Gender (Def)                  & 0                   & 2             \\ \cline{2-4} 
                        & \multirow{5}{*}{Gender}       & 1                   & 249           \\
                        &                               & 3                   & 247           \\
                        &                               & 5                   & 249           \\
                        &                               & 10                  & 250           \\
                        &                               & 20                  & 13            \\ \cline{2-4} 
                        & Gen \& Conti (Def)            & 0                   & 8             \\ \cline{2-4} 
                        & \multirow{5}{*}{Gen \& Conti} & 1                   & 246           \\
                        &                               & 3                   & 248           \\
                        &                               & 5                   & 249           \\
                        &                               & 10                  & 171           \\
                        &                               & 20                  & 6             \\ \hline
\end{tabular}%
}
\caption{Number of examples across all experiment settings.}
\label{tab:freq_dist_examples}
\end{table*}

\subsection{Prompts}
\begin{tcolorbox}[title={Prompt}, width=\textwidth, colback=white, colframe=gray, arc=0pt, outer arc=5pt, boxrule=0.5pt, leftrule=2pt, rightrule=2pt, right=0pt, left=0pt, top=0pt, bottom=0pt, toprule=0pt, bottomrule=2pt]
\label{fig:sample_prompt}
% \small

\# AI Rules\\
- Output response as a Python list only.\\
- Do not output any extra text.\\
- Do not wrap the response in Python markers.\\
- Do not assign the list to any variable.\\
- List values in double-quotes.\\
    
You are proficient in recommending new movie for users to watch based on their background, previous view history, or a combination of both. \\
The user is a 25-34 years old Male clerical/admin.\\
The user has previously watched the following movies: ['Out of Sight (1998)', 'Horse Whisperer, The (1998)', 'Star Wars: Episode V - The Empire Strikes Back (1980)', 'Odd Couple II, The (1998)', 'Marathon Man (1976)'].\\
From the candidates listed below, recommend the next 10 movie for the user to watch based on the user's background, previous view history, or a combination of both.\\
Format your response as a Python list of item names. The list must be ranked from the most likely to the least likely movie.\\
Candidates: ['Mr. \& Mrs. Smith (1941)', 'Blue Velvet (1986)', 'Freedom for Us (À nous la liberté ) (1931)', 'White Balloon, The (Badkonake Sefid ) (1995)', 'Fear, The (1995)', 'Barefoot Executive, The (1971)', 'Barb Wire (1996)', 'Jungle Book, The (1967)', 'Matrix, The (1999)', '24-hour Woman (1998)', 'Heat (1995)', 'Fish Called Wanda, A (1988)', 'Independence Day (ID4) (1996)', 'Dead Calm (1989)', 'Phantasm III: Lord of the Dead (1994)', 'To Be or Not to Be (1942)', 'Rain Man (1988)', 'Carnosaur (1993)', 'Heathers (1989)', 'Gaslight (1944)', 'Get Bruce (1999)', 'Omen, The (1976)', 'Bedknobs and Broomsticks (1971)', 'Herbie Rides Again (1974)', 'Buck and the Preacher (1972)', 'Wallace \& Gromit: The Best of Aardman Animation (1996)', 'Friday the 13th Part 3: 3D (1982)', 'Meatballs (1979)', 'Cabin Boy (1994)', '8 Heads in a Duffel Bag (1997)', 'Mariachi, El (1992)', 'Contender, The (2000)', 'When a Man Loves a Woman (1994)', 'Henry Fool (1997)', 'Beetlejuice (1988)', 'Requiem for a Dream (2000)', 'Raven, The (1963)', 'Grand Day Out, A (1992)', 'Miami Rhapsody (1995)', 'Tales From the Crypt Presents: Demon Knight (1995)', 'Sticky Fingers of Time, The (1997)', 'Opposite of Sex, The (1998)', 'Saving Private Ryan (1998)', 'Naked Gun 33 1/3: The Final Insult (1994)', 'Trigger Effect, The (1996)', 'Among Giants (1998)', 'Spaceballs (1987)', 'Bloody Child, The (1996)', 'Snow White and the Seven Dwarfs (1937)', 'Man Who Knew Too Much, The (1934)']\\

\end{tcolorbox}

\subsection{Additional Plots}
\label{additional_plots}

\begin{figure*}[!t]
    \centering 
    \includegraphics[width=\linewidth]{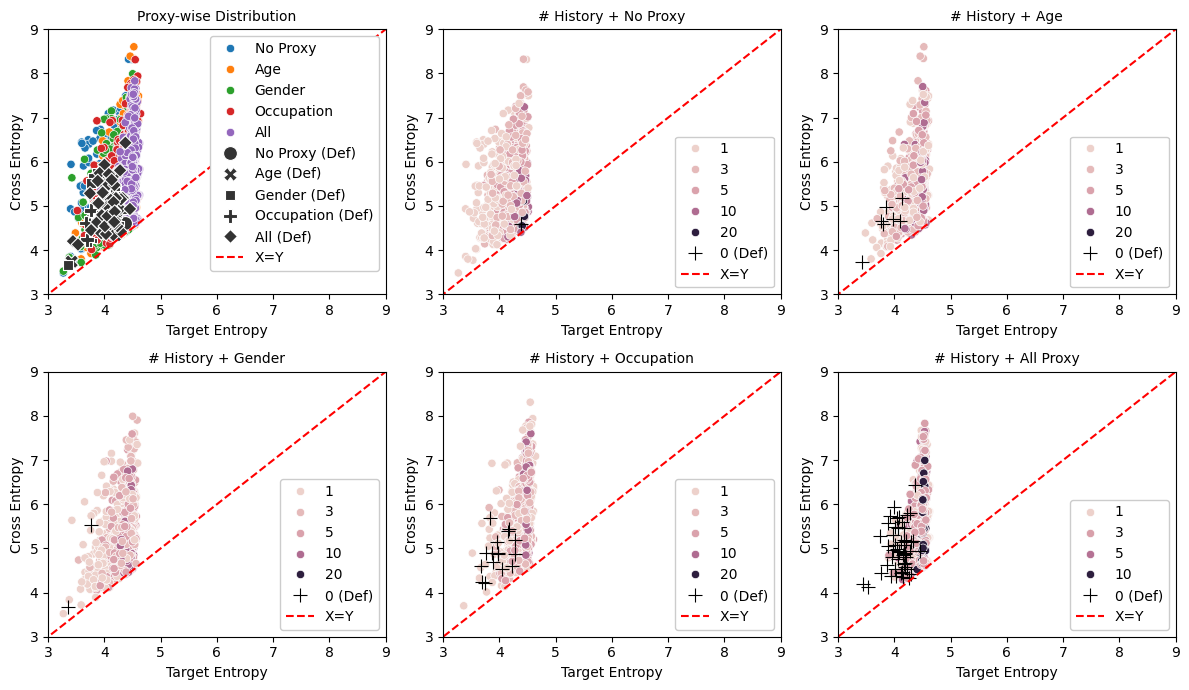} 
    \caption{GPT-4o detailed plot for Movie}
    \label{fig:d1}
\end{figure*}

\begin{figure*}[!t]
    \centering 
    \includegraphics[width=\linewidth]{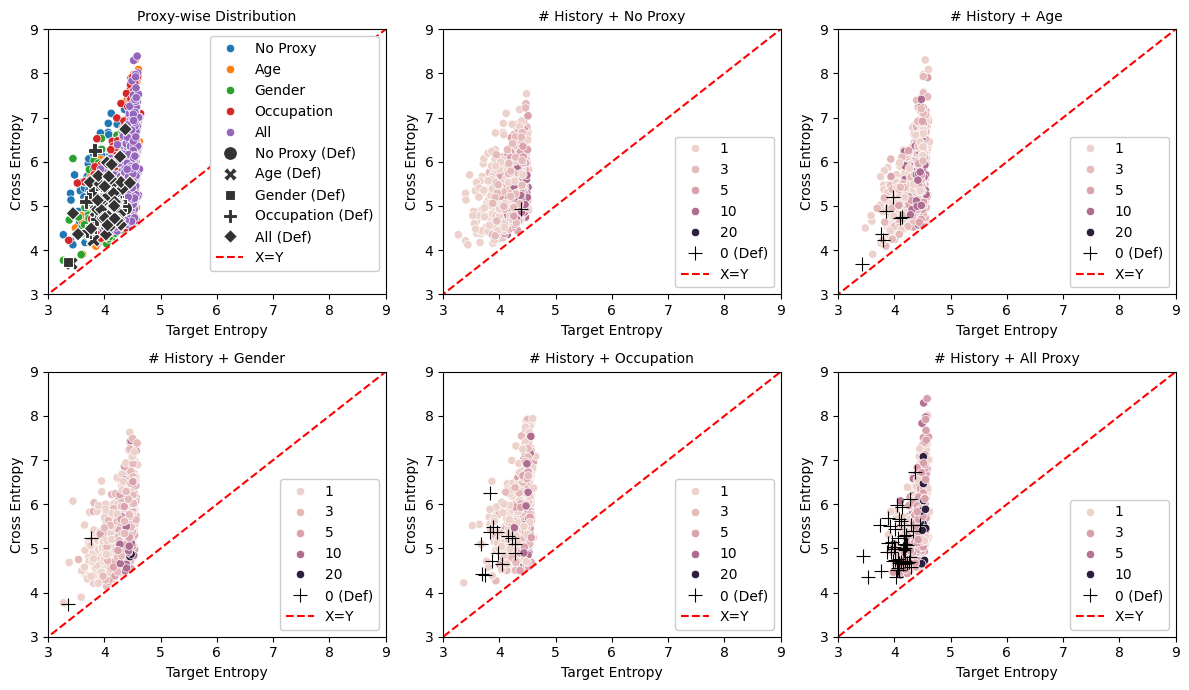} 
    \caption{GPT-4o-mini detailed plot for Movie}
    \label{fig:d2}
\end{figure*}

\begin{figure*}[!t]
    \centering 
    \includegraphics[width=\linewidth]{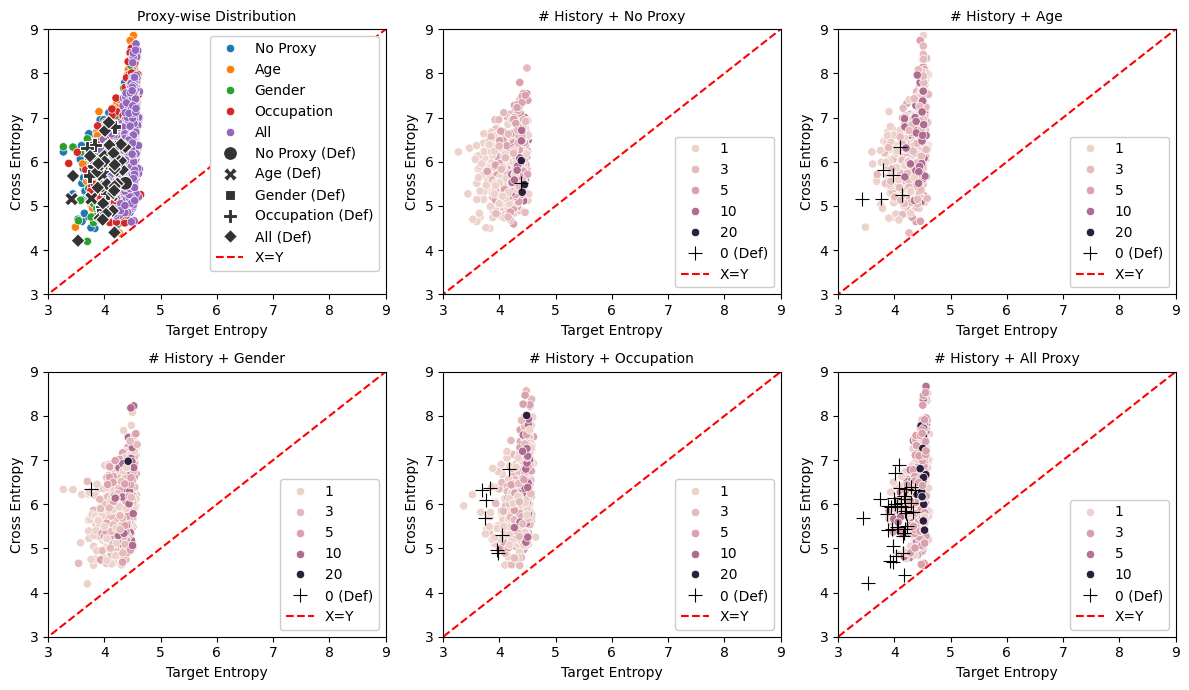} 
    \caption{Llama detailed plot for Movie}
    \label{fig:d7}
\end{figure*}

\begin{figure*}[!t]
    \centering 
    \includegraphics[width=\linewidth]{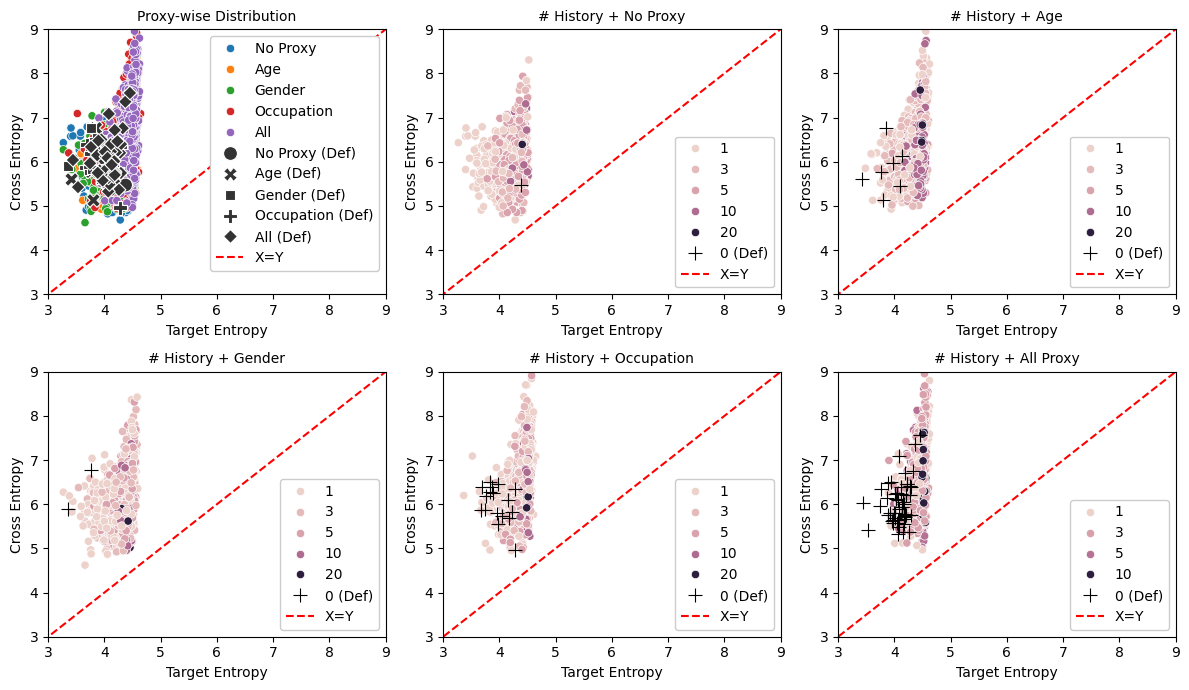} 
    \caption{Random detailed plot for Movie}
    \label{fig:d3}
\end{figure*}

\begin{figure*}[!t]
    \centering 
    \includegraphics[width=\linewidth]{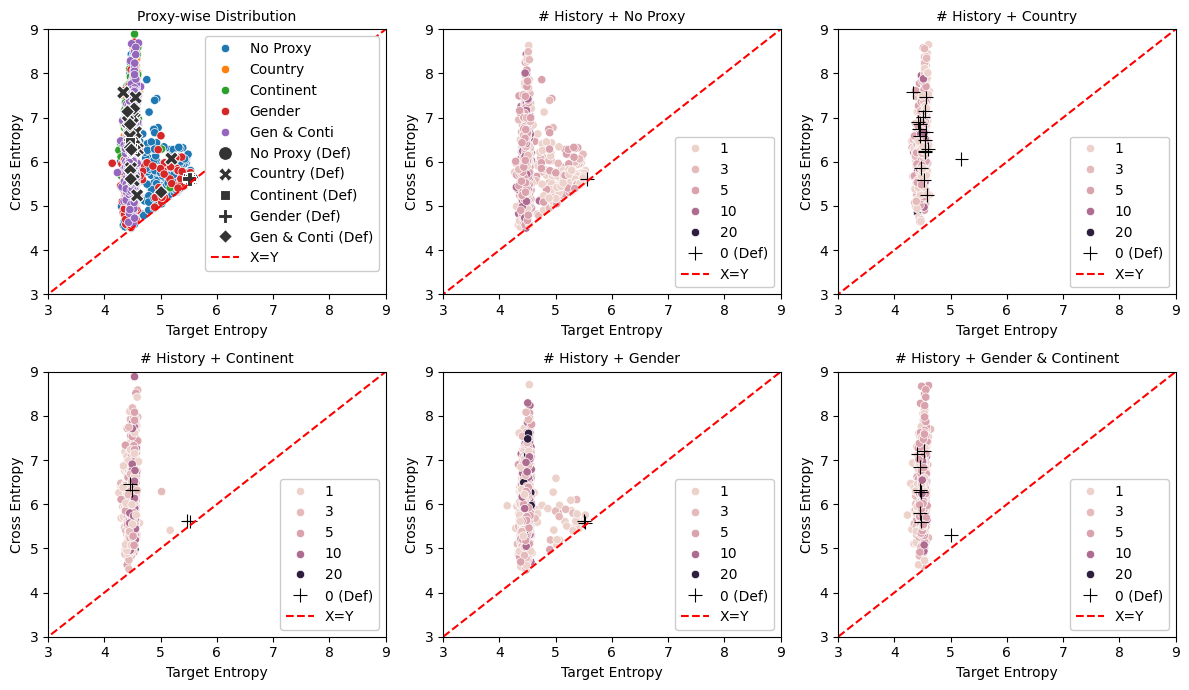} 
    \caption{GPT-4o detailed plot for Music}
    \label{fig:d4}
\end{figure*}

\begin{figure*}[!t]
    \centering 
    \includegraphics[width=\linewidth]{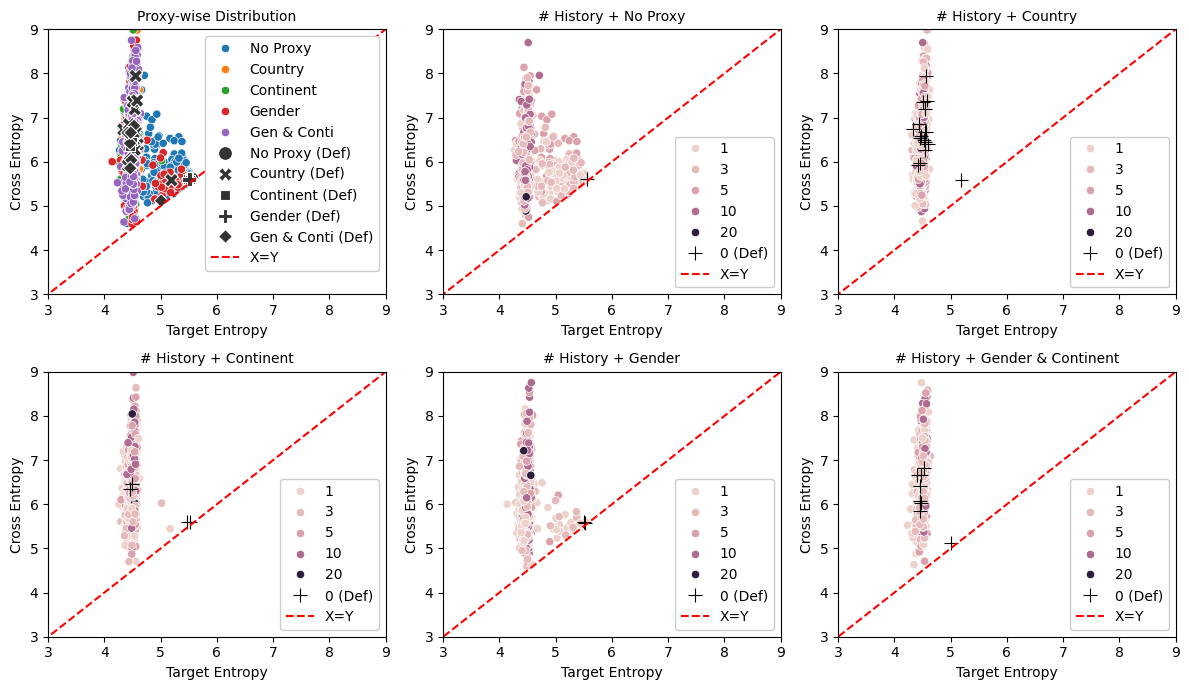} 
    \caption{GPT-4o-mini detailed plot for Music}
    \label{fig:d5}
\end{figure*}

\begin{figure*}[!t]
    \centering 
    \includegraphics[width=\linewidth]{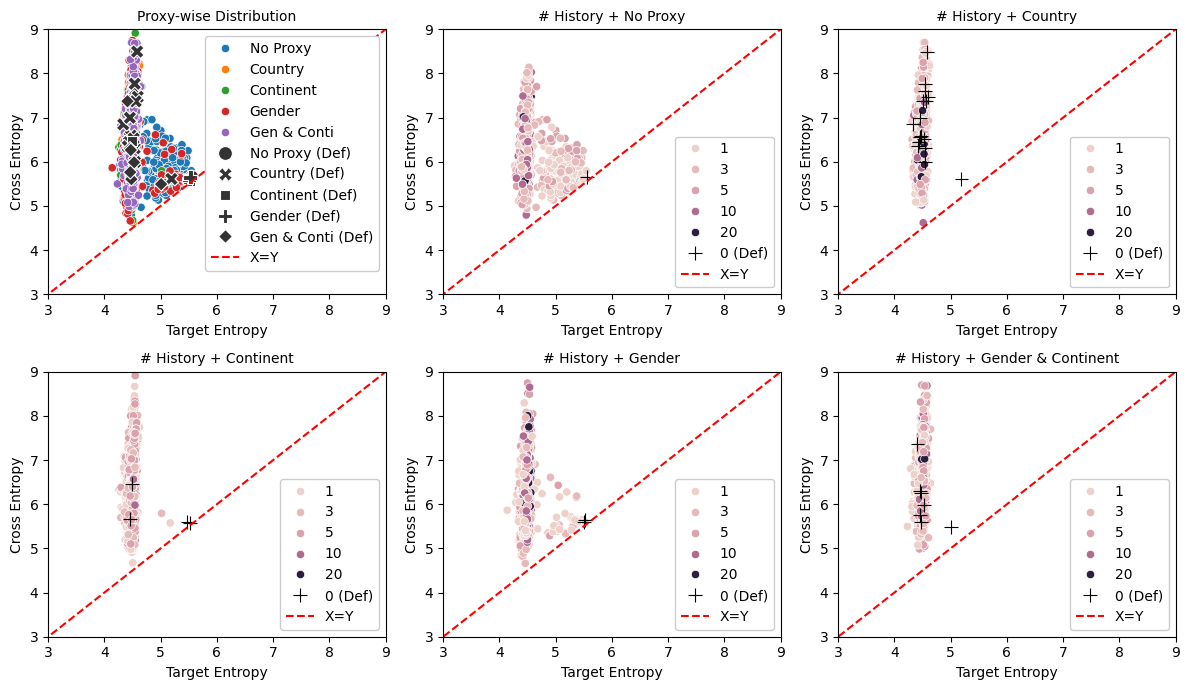} 
    \caption{Llama detailed plot for Music}
    \label{fig:d8}
\end{figure*}

\begin{figure*}[!t]
    \centering 
    \includegraphics[width=\linewidth]{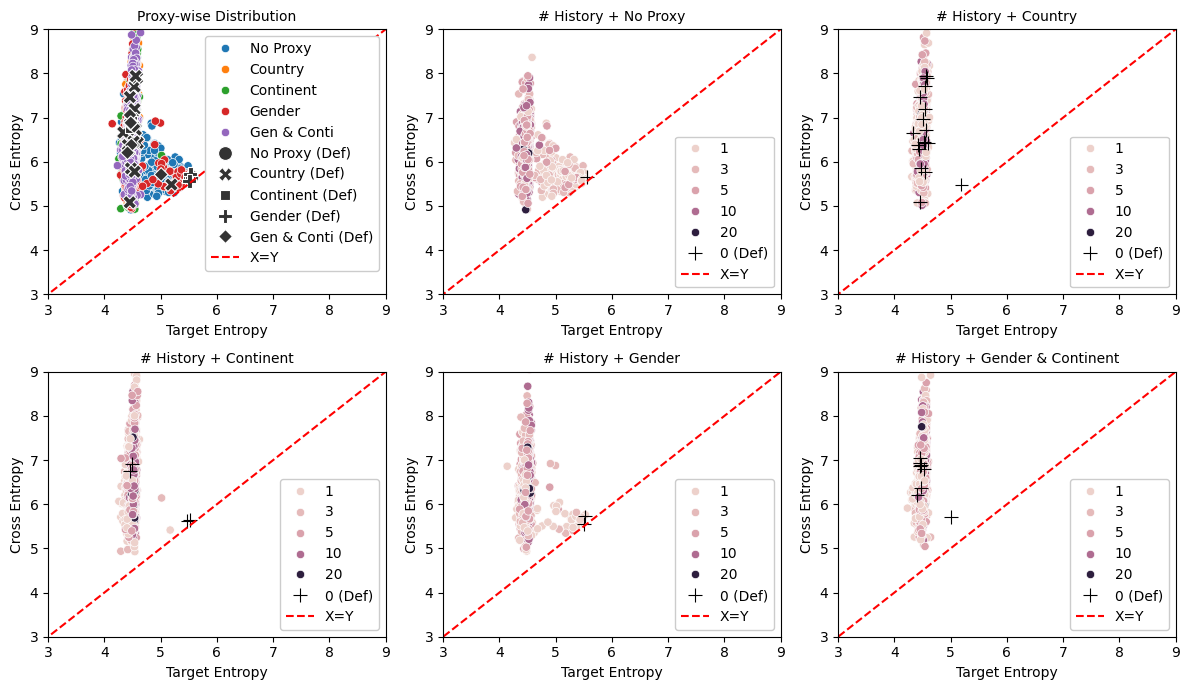} 
    \caption{Random detailed plot for Music}
    \label{fig:d6}
\end{figure*}

\end{document}